\documentclass[10pt,twocolumn,letterpaper]{article}

\usepackage{iccv}
\usepackage{times}
\usepackage{epsfig}
\usepackage{graphicx}
\usepackage{amsmath}
\usepackage{amssymb}

\newcommand{\argmin}{\operatornamewithlimits{argmin}}

\newcommand{\vect}[1]{\boldsymbol{\mathbf{#1}}}

\usepackage{multirow}
\usepackage{multicol}
\usepackage[ruled,vlined]{algorithm2e} 

\usepackage[breaklinks=true,bookmarks=false]{hyperref}

\iccvfinalcopy 

\def\iccvPaperID{****} 

\ificcvfinal\fi

\begin{document}

\title{A Simple Approach for Zero-Shot Learning based on \\Triplet Distribution Embeddings}

\author{Vivek Chalumuri\\
KTH Royal Institute of Technology\\
Stockholm, Sweden\\
{\tt\small vivekc@kth.se}
\and
Bac Nguyen\\
Sony Europe B.V.\\
Stuttgart Laboratory 1, Germany\\
{\tt\small Bac.NguyenCong@sony.com}
}

\maketitle
\ificcvfinal\thispagestyle{empty}\fi

\begin{abstract}
Given the semantic descriptions of classes, Zero-Shot Learning (ZSL) aims to recognize unseen classes without labeled training data by exploiting semantic information, which contains knowledge between seen and unseen classes. Existing ZSL methods mainly use vectors to represent the embeddings to the semantic space. Despite the popularity, such vector representation limits the expressivity in terms of modeling the intra-class variability for each class. We address this issue by leveraging the use of distribution embeddings. More specifically, both image embeddings and class embeddings are modeled as Gaussian distributions, where their similarity relationships are preserved through the use of triplet constraints. The key intuition which guides our approach is that for each image, the embedding of the correct class label should be closer than that of any other class label. Extensive experiments on multiple benchmark data sets show that the proposed method achieves highly competitive results for both traditional ZSL and more challenging Generalized Zero-Shot Learning (GZSL) settings.
\end{abstract}

\section{Introduction}
The availability of large-scale data has led to significant advances in computer vision over the last decades~\cite{simonyan2014very, szegedy2015going, he2016deep}. In a conventional image classification task, a classifier is trained to discriminate images among various classes. Because classifiers are trained to separate a deterministic set of classes, they often fail to classify images from classes that are unavailable during the training. Humans, on the other hand, are extremely good at recognizing novel visual classes by leveraging over the existing knowledge about the classes (\emph{e.g.}, description of the classes). For instance, given the information that \textit{zebras look similar to horses but with black and white stripes}, having seen a horse one may identify a zebra without ever seeing it before. Recognizing novel visual categories without labeled training examples is also known as Zero-Shot Learning (ZSL)~\cite{wang2019survey,Elhoseiny_2019_ICCV,Kato_2019_ICCV_Workshops}. In addition to labeled training data, auxiliary semantic information that relates unseen classes to seen classes are provided. The type of semantic information or high-level description depends on data sets, for instance, human-annotated class attributes~\cite{lampert2013attribute}, word vectors~\cite{akata2015evaluation}, or natural language descriptions~\cite{zhu2018generative} of the class labels.

\begin{figure}[t]
    \centering
    \includegraphics[scale=0.5]{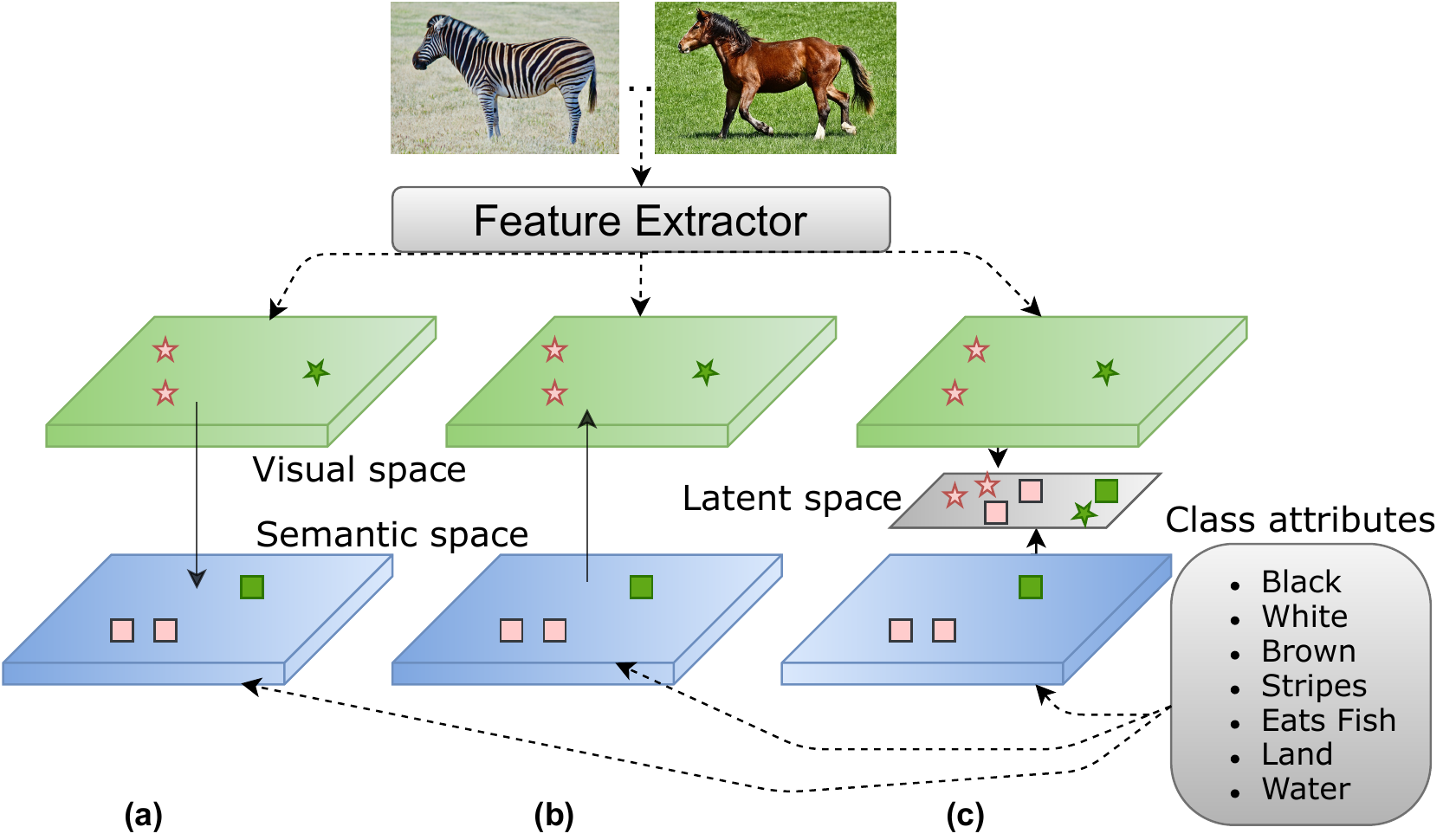}
  \caption{Illustration of common embedding-based ZSL methods: (a) learning a mapping from the visual space to the semantic space, (b) learning a mapping from the semantic space to the visual space, and (c) learning two mappings from the visual space and the semantic space to a common latent space.}
  \label{fig:ZSL_Methods}
\end{figure}

Despite the fact that seen and unseen class labels are disjoint sets, they are related by the semantic descriptions. A simple idea is to learn a semantic mapping so that representations of both seen and unseen examples share the same semantic space. Common ZSL embedding-based methods can be roughly categorized into three cases as illustrated in Fig.~\ref{fig:ZSL_Methods}. Early approaches~\cite{palatucci2009zero, frome2013devise, akata2015evaluation} learned a mapping from the visual space to the semantic space and then perform a nearest-neighbor search to find the correct label (see Fig.~\ref{fig:ZSL_Methods}a). An opposite direction~\cite{kumar2018generalized,xian2018feature} is to learn a mapping from the semantic space to the visual space (see Fig.~\ref{fig:ZSL_Methods}b). Methods of this category treat ZSL as a missing data problem and try to synthesize unseen class image features from the projections of the corresponding semantic information. Other approaches~\cite{hubert2017learning,changpinyo2016synthesized} learn mappings from both visual and semantic information into some common latent space where they share semantically-meaningful information (see Fig.~\ref{fig:ZSL_Methods}c). Despite the initial success, most of these approaches represent the class labels and image features as vector embeddings. Since vector embeddings are point estimates, they cannot capture the intra-class variability among different classes. A possible solution to overcome this limitation is to use distribution embeddings instead of the usual vector embeddings. More specifically, given a data point, we learn a function that maps it into a distribution (\textit{e.g}., multivariate Gaussians) over the possible values from which this data point might have been generated. Therefore, the intra-class variability can be better modeled. In our method, both image features and class attributes are modeled as distributions in order to further increase the flexibility of the model.

There have been a few attempts following this direction~\cite{verma2017simple, mishra2018generative, schonfeld2019generalized}. The most relevant study comes from Schonfield \etal~\cite{schonfeld2019generalized} with CADA-VAE, an encoder-decoder paradigm to construct visual and semantic distributions in a latent space. More specifically, the distribution embeddings are aligned by forcing a cross-modal reconstruction criterion, which enables the knowledge transfer between seen and unseen classes. Unlike CADA-VAE, we adopt a simple triplet loss function to learn distribution embeddings. Its objective requires the visual distribution embedding of an image being closer to the semantic distribution embedding of its ground-truth class than that of other classes. Our method has the advantage of the simplicity of implementation while achieving better results than CADA-VAE.


The main contributions of the paper can be summarized as follows. (1) We introduce a simple yet effective idea that considers ZSL as learning distribution embeddings. Our method follows an end-to-end learning fashion, which can potentially lead to a better embedding space. More specifically, we directly optimize a loss function that makes the distribution of each image being close to the distribution of its class label, while being far away from those of other class labels. By learning from the similarity (distance) instead of category-specific concepts, our method can naturally deal with problems involving unseen classes. (2) To make learning more effective and efficient, we design an online strategy of building triplet distribution constraints. (3) Extensive experiments on public data sets are carried out to validate the effectiveness of our method in both the ZSL and Generalized Zero-Shot Learning (GZSL) settings.


\section{Related Work}
Recently, ZSL has gained growing interest in computer vision  due to lack of labeled data. In many practical applications, there may not be training examples for classes appearing in the test set. Inspired by the way human recognizing the visual images, a number of ZSL approaches have been proposed to deal with these problems using some auxiliary semantic information~\cite{lampert2013attribute,lampert2009learning,kankuekul2012online}. The semantic information can be obtained from Word2Vec~\cite{akata2015evaluation}, Glove~\cite{pennington2014glove}, WordNet~\cite{miller1995wordnet}, or  natural language descriptions~\cite{reed2016learning,li2019zero}. More recently, Karessli \etal~\cite{karessli2017gaze} have exploited the use of human gaze localization. Regardless of the way that semantic information is obtained, the crux of ZSL methods is to learn a mapping between the visual space and the semantic space using only the information from the seen classes. 


Earliest works~\cite{lampert2009learning,palatucci2009zero,norouzi2013zero} focused directly or indirectly on mapping image features into the semantic space. Once the mapping is learned, image features can be projected into the semantic space and the prediction is based on a nearest-neighbor search over the unseen class-attribute embeddings. As an extension, Bucher \etal~\cite{bucher2016improving} formulated ZSL as a distance metric learning problem, where image features are projected into the semantic space. Thus, a linear mapping induced by the Mahalanobis distance metric is learned to capture the similarity relations between image features and class attributes.


Another predominant direction is to learn a bilinear compatibility function between the visual space and the semantic space. Essentially, a bilinear compatibility function should make objects from the same class being closer to each other and objects from different classes being far away from each other~\cite{xian2016latent}. Different loss functions have been used to learn such a bilinear function. For instances, ALE~\cite{akata2015label} and DeVise~\cite{frome2013devise} used a ranking loss function, while SJE~\cite{akata2015evaluation} used the structured support vector machine (SVM) loss. To avoid the limitation of parametric bilinear functions, Zhang \etal~\cite{zhang2017learning} computed the compatibility score between visual and semantic features through a deep neural network. Following this end-to-end learning fashion, we employ two neural networks, which directly maximize the similarity between distribution embeddings in order to learn meaningful representation.

\begin{figure*}
    \centering
    \includegraphics[width=0.9\textwidth]{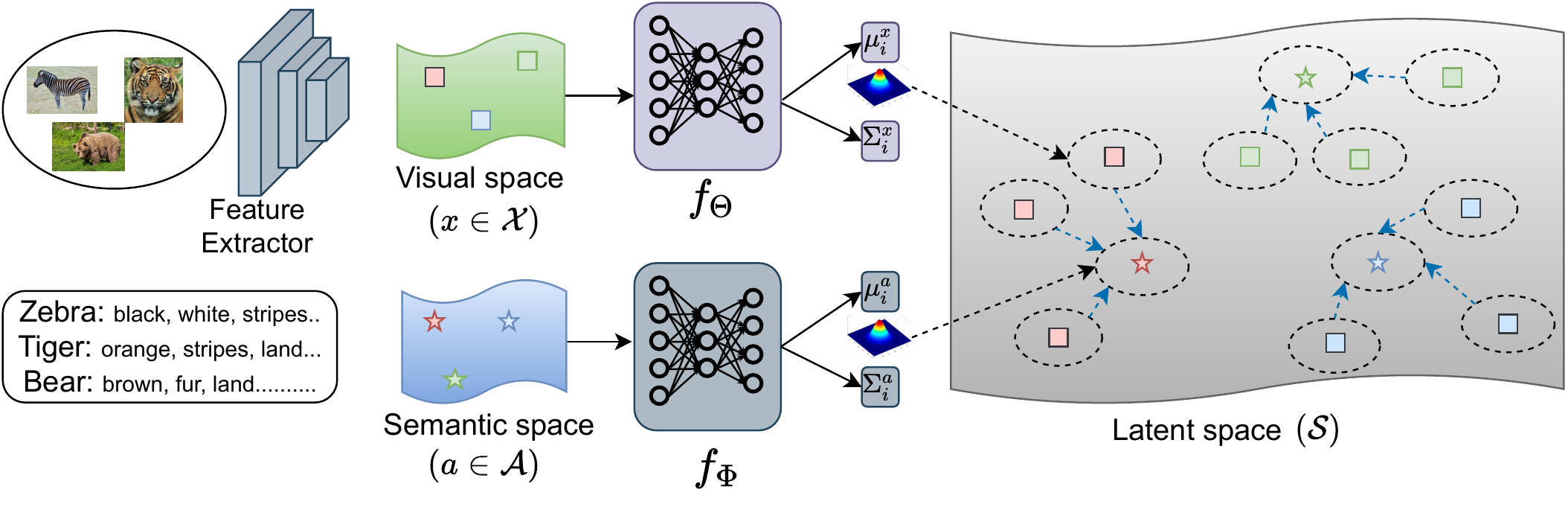}
  \caption{An overview of the proposed method. The dotted ellipses in the latent space represent the distribution embeddings. The distribution embeddings of image features and class attributes are distinguished by square and pentagram symbols at the center, respectively. Examples of the same class are denoted with same color. Note that Feature Extractor is not part of our framework.}
  \label{fig:TriModel}
\end{figure*}


A key drawback of many ZSL approaches is that vector embeddings are relatively limited in their expressivity. Such vector embeddings lead to the loss of information regarding the intra-class variability of each class. To alleviate this issue, Mukherjee and Hospedales~\cite{mukherjee2016gaussian} proposed the use distribution embeddings for both image and semantic features. More specifically, they represented images and class labels as Gaussian distributions and learned a multi-modal mapping between two spaces, which is parameterized by a simple linear transformation matrix. Since distribution embeddings are learned independently, it may lead to an suboptimal solution. Similarly, Verman and Rai~\cite{verma2017simple} modeled each class-conditional distribution as an exponential family distribution and obtained better results compared to the earlier methods. However, they used a two-phase approach to first learn the distribution parameters for the seen classes using maximum likelihood estimators, then estimate a linear model that associates between class attributes and distributions for the unseen classes. Unlike these methods, we focus on modeling both image features and class attributes as distributions using deep neural networks in an end-to-end learning fashion, which can lead to a better embedding space.

Another way of looking at ZSL is by formulating it as a missing data problem, where examples from the unseen classes are missing. A few generative frameworks for ZSL were proposed~\cite{mishra2018generative,chen2018zero,schonfeld2019generalized,yu2020episode}. For instance, Chen \etal~\cite{chen2018zero} presented an adversarial learning framework, which can generate raw images. Mishra \etal~\cite{mishra2018generative} used conditional variational autoencoders to generate image features.  On the other hand, Schonfeld \etal~\cite{schonfeld2019generalized} applied the aligned variational autoencoders to generate image features in a low-dimensional latent space. Having also examples generated from the unseen classes, ZSL can now be treated as a supervised learning problem, which can be solved using a conventional classifier. By taking into account examples generated from the unseen classes, we can alleviate the model bias problem in ZSL since only examples from the seen classes are involved during training. Our method naturally extends this idea as we learn distribution embeddings. More recently, Yunlong \etal~\cite{yu2020episode} in their generative approach used a complicated episodic training to synthesize class-level visual prototypes conditioned on semantic information instead of synthesizing instance-level features. Contrary to this, our approach is much more straightforward in the sense that we learn distribution embeddings directly for both visual features and semantic prototypes with a simple training framework employing triplet constraints.




\section{Methodology}

We start by formulating the ZSL problem in a formal way. Let $\mathcal{Y}^{s}$ and $\mathcal{Y}^{u}$ denote the sets of seen and unseen class labels, respectively. Note that $\mathcal{Y}^{s}$ and $\mathcal{Y}^{u}$ are disjoint sets, \emph{i.e.}, $\mathcal{Y}^{s} \cap \mathcal{Y}^{u} = \emptyset$. A set of $N$ training examples is defined as $\mathcal{D}^{tr} = \{(\vect{x}_i, y_i)\}_{i=1}^{N}$, where $\vect{x}_i\in \mathcal{X} \subset  \mathbb{R}^D$ denotes the $i$-th example (\emph{e.g.}, image) with its corresponding class label $y_i\in \mathcal{Y}^s$. In addition, each class label $y_i$ is associated with a semantic description (\emph{e.g.}, class-attribute vector) $\vect{a}_i\in \mathcal{A} \subset \mathbb{R}^L$. The goal of ZSL is to learn a classifier $f\colon \mathcal{X} \to \mathcal{Y}^u$ for an unseen example $\vect{x}_t$. In the GZSL setting, the unseen example may be assigned to either a seen or unseen class, \emph{i.e.}, $y_t \in \mathcal{Y}^{s} \cup \mathcal{Y}^{u}$.


\subsection{Proposed Approach}
We aim to learn an effective embedding space in which image features and class attributes are represented as distributions. Our goal is to discover such a representation where correlations between images and labels are directly modeled as pairwise similarity. This can be formulated within a large-margin framework, which ensures that for each image the correct label is closer than any other label. In other words, images of the same class should be close to the corresponding class attributes, whereas images of highly different classes should be well separated. Classification tasks will therefore be reduced to the nearest-neighbor problem in the embedded space, leveraging the fact that nearest neighbors have been preserved during training. An overview of our method is illustrated in Fig.~\ref{fig:TriModel}.


As the first step, the visual and semantic representations are embedded as distributions in the latent space. To this end, we use two separate neural networks that transform an input vector into a distribution. For simplicity, we assume that embeddings are multivariate Gaussian distributions with diagonal covariance matrices. Each neural network performs in a different space. More formally, we consider a differentiable function  $f_{\vect{\Theta}}\colon \mathcal{X}\to \mathcal{S}$ parameterized by $\vect{\Theta}$, which maps an image $\vect{x}_i$ from the visual space into the latent space $\mathcal{S}$. For the visual space, the image feature $\vect{x}_i$ is computed by feeding an image into a pre-trained deep convolutional neural network. Similarly, let $f_{\vect{\Phi}}\colon \mathcal{A} \to \mathcal{S}$ denote a differentiable function that maps an attribute vector $\vect{a}_i$  from the semantic space into the latent space $\mathcal{S}$. Here, both $f_{\vect{\Theta}}$ and $f_{\vect{\Phi}}$ are defined as multi-layer perceptrons (MLPs). Each distribution in $\mathcal{S}$ is parameterized by a mean vector $\vect{\mu}_{i} \in \mathbb{R}^K$ and a diagonal covariance matrix $\vect{\Sigma}_i\in \mathbb{R}^{K\times K}$, whose diagonal elements are non-negative.

In this paper, we focus on the Wasserstein distance, also known as the Earth Mover’s distance, to measure the dissimilarity between two distributions. The $p$-Wasserstein distance between two probability distributions $\mathbb{P}$ and $\mathbb{Q}$ is defined as
\begin{align}
    d(\mathbb{P}, \mathbb{Q}) = \left(\inf_{\gamma \sim \Pi(\mathbb{P}, \mathbb{Q})} \mathbb{E}_{(\vect{u}, \vect{v}) \sim \gamma}[\| \vect{u}-\vect{v} \|^p]\right)^{1/p}\,,
    \label{Eq:Wasserstein}
\end{align}
where $\Pi(\mathbb{P}, \mathbb{Q})$ denotes the set of all possible joint probability distributions whose marginals are $\mathbb{P}$ and $\mathbb{Q}$, respectively. Roughly speaking, it measures the amount of ``mass'' we must transport from $\vect{u}$ to $\vect{v}$ in order to transform from the distribution $\mathbb{P}$ into the distribution $\mathbb{Q}$ with an optimal transport plan. Early works~\cite{pmlr-v70-arjovsky17a,10.5555/3295222.3295327} have shown very promising results in measuring distances between two distributions with the Wasserstein distance metric, which often produces more stable gradients even if two distributions are not overlapped. Despite its nice properties, the infimum in Eq.~(\ref{Eq:Wasserstein}) is highly intractable, making it inapplicable for arbitrary distributions. Fortunately, for multivariate Gaussian distributions as in our case, the 2-Wasserstein distance metric admits a closed-form solution~\cite{givens1984class}, given by
\begin{equation}
\begin{aligned}
    &d^2\big(\mathcal{N}(\vect{\mu}_1, \vect{\Sigma}_1),\mathcal{N}(\vect{\mu}_2, \vect{\Sigma}_2)\big)\\
    &= \|\vect{\mu}_1-\vect{\mu}_2\|_2^2 
    + \mathrm{tr}(\vect{\Sigma}_1) + \mathrm{tr}(\vect{\Sigma}_2) - 2(\vect{\Sigma}_2^\frac{1}{2}\vect{\Sigma}_1\vect{\Sigma}_2^\frac{1}{2})^\frac{1}{2}\,.
\end{aligned}
\end{equation}

Once distribution embeddings are obtained, we define a loss function on top of the latent space. Ideally, the mappings $f_{\vect{\Phi}}$ and $f_{\vect{\Theta}}$ should have the following properties
\begin{equation}
\begin{aligned}
    d(f_{\vect{\Theta}}(\vect{x}_i), f_{\vect{\Phi}}(\vect{a}_i)) < d(f_{\vect{\Theta}}(\vect{x}_i), f_{\vect{\Phi}}(\vect{a}_j)), \\
    \forall (\vect{x}_i, \vect{a}_i, \vect{a}_j) \text{ for which } y_i \ne y_j\,. \label{eq:triplet}
\end{aligned}
\end{equation}
Here, $\vect{a}_i$ is called a positive-class neighbor and $\vect{a}_j$ a negative-class neighbor of~$\vect{x}_i$. Apparently, the number of possible triplet constraints $(\vect{x}_i, \vect{a}_i, \vect{a}_j)$ in~(\ref{eq:triplet}) can be huge. We will introduce an efficient strategy to overcome this issue in the next subsection. To simplify the mathematical expression, let denote a set of triplet constraints $(\vect{x}_i, \vect{a}_i, \vect{a}_j)$ as
\begin{align}
    \mathcal{C} = \{(\vect{x}_i, \vect{a}_i, \vect{a}_j) \mid \vect{x}_i \text{ should be closer to } \vect{a}_i \text{ than to } \vect{a}_j\}\,.
\end{align}
In order to enforce the triplet constraints in $\mathcal{C}$, we jointly learn the mappings $f_{\vect{\Phi}}$ and $f_{\vect{\Theta}}$ by solving following optimization problem
\begin{equation}
    \min_{\vect{\Theta}, \vect{\Phi}} \,\,\!\!\frac{1}{|\mathcal{C}|}\!\!\!\!\!\!\sum_{{(\vect{x}_i, \vect{a}_i, \vect{a}_j)\in \mathcal{C}}}\!\!\!\!\!\!\!\!\Big[d(f_{\vect{\Theta}}(\vect{x}_i), f_{\vect{\Phi}}(\vect{a}_i))- d(f_{\vect{\Theta}}(\vect{x}_i), f_{\vect{\Phi}}(\vect{a}_j))+\epsilon\Big]_{+} \label{eq:obj}
\end{equation}
with $\epsilon >0$ a predefined margin and $[.]_{+}=\max(., 0)$ the hinge loss function that penalizes the violated constraints. This loss function ensures that the image embedding is closer to its correct label embedding than to any other incorrect label embedding by at least a margin $\epsilon$. Since the loss function in~(\ref{eq:obj}) is differentiable, we can adopt the back-propagation algorithm~\cite{kelley1960gradient} to compute the gradients and employ stochastic gradient descent to solve it. We refer to our method as \textit{Distribution Embeddings based on Triplet Constraints} (DETC).


It is important to note that images of different classes will be pushed far away from each other in an implicit way. This is due to the fact that each label embedding is only surrounded by examples belonging to its class. Therefore, examples from different classes will not be clustered together. This is beneficial for the ZSL tasks in order to generalize well to unseen classes.

Given the learned parameters $\vect{\Theta}$ and $\vect{\Phi}$, the prediction for an unseen example $\vect{x}_t$ can be employed as
\begin{align}
    \hat{y}_t = \argmin_{y \in \mathcal{Y}^u} d(f_{\vect{\Theta}}(\vect{x}_t), f_{\vect{\Phi}}(\vect{a}))\,,
\end{align}
where $\vect{a}$ denotes the corresponding class-attribute vector with the label $y$. Differently from \cite{9051798}, we assume that our MLPs parameterized by $\vect{\Theta}$ and $\vect{\Phi}$ are capable of modelling any complex function that produces meaningful representations. Therefore, there is no need to learn an auxiliary classifier since a simple nearest-neighbor classifier can perform very well on the latent space.


\subsection{Triplet Constraints}
Another main challenge is to collect a set of meaningful constraints $\mathcal{C}$ in problem~(\ref{eq:obj}). For an image $\vect{x}_i$, we can construct $O(|\mathcal{Y}^s|)$ triplet constraints by simply forming $(\vect{x}_i, \vect{a}_i, \vect{a}_j)$, where $y_i \ne y_j$. However, using all possible triplet constraints can result in very inefficient training, especially when the number of class labels is high. One may use a random sampling strategy to select a subset of constraints instead of all possible constraints. Nevertheless, this strategy can induce easy constraints, which produce gradients with zero or small magnitudes~\cite{NGUYEN2020209}. Consequently, they contribute little to the gradient computation. Therefore, stochastic gradient descent can suffer from slow convergence. 


To reduce the number of triplet constraints, an online triplet-sampling strategy is proposed as follows. For each image $\vect{x}_i$, we only select one negative-class neighbor $\vect{a}^{-}_{i}$ defined as the nearest class embedding with a different class label, \emph{i.e.},
\begin{align}
\vect{a}^{-}_{i} = \argmin_{\vect{a}_j \mid y_j \in \mathcal{Y}^s\setminus \{y_i\}} \,d( f_{\vect{\Theta}}(\vect{x}_i), f_{\vect{\Phi}}(\vect{a}_j))\,.
\label{eq:hardtriplet}
\end{align}
By doing so, we reduce the number of possible triplet constraints for each $\vect{x}_i$ from $O(|\mathcal{Y}^s|)$ to $O(1)$. Since we employ stochastic gradient descent with a mini-batch $\mathcal{B}$ sampled from the feature space to solve problem~(\ref{eq:obj}), the number of triplet constraints on each iteration is greatly reduced from $O(|\mathcal{Y}^s|\,|\mathcal{B}|)$ to $O(|\mathcal{B}|)$. Note that the negative-class neighbors may change during the learning process, depending on the current values of the parameters $\vect{\Theta}$ and $\vect{\Phi}$. Therefore, we recompute the negative-class neighbors on each mini-batch update in an online manner. Finally, problem~(\ref{eq:obj}) can be approximated as
\begin{align}
     \min_{\vect{\Theta}, \vect{\Phi}} &\, \frac{1}{|\mathcal{B}|}
     \sum_{\vect{x}_i\in\mathcal{B}}\Big[d(f_{\vect{\Theta}}(\vect{x}_i), f_{\vect{\Phi}}(\vect{a}_i))\!-\! d(f_{\vect{\Theta}}(\vect{x}_i), f_{\vect{\Phi}}(\vect{a}^{-}_{i}))\!+\!\epsilon\Big]_{+}\!\!\!. \label{eq:batchloss}
\end{align}
The pseudocode for the proposed method is given in Algorithm~\ref{algo:training}.


\begin{algorithm}[t]
\SetAlgoLined
\SetKwInOut{Input}{Input}
\SetKwInOut{Output}{Output}
\Input{A training set $\mathcal{D}^{tr}=\{(\vect{x}_i,y_i)\}_{i=1}^N$, and its associated class-attribute vectors $\{\vect{a}_i\}_{i=1}^N$}
 Initialize $\vect{\Theta}, \vect{\Phi}$ \;
\For{$r=1,2,....,T$}{
i) Randomly sample a mini-batch of examples $\mathcal{B}$\\
 ii) Project $\vect{x}_i \in \mathcal{B}$ and its corresponding $\vect{a}_i$ into the latent space\\
 iii) Compute the negative-class neighbor for each $\vect{x}_i$ using Eq.~(\ref{eq:hardtriplet}). \\
 iv) Update parameters $\vect{\Theta}$ and $\vect{\Phi}$ by solving problem~(\ref{eq:batchloss}).
 }
 \Output{Model parameters $\vect{\Theta}$ and $\vect{\Phi}$}
 \caption{Training DETC}\label{algo:training}
\end{algorithm}

\subsection{Data Generation}
A major advantage of having distribution embeddings is that we can generate examples of unseen classes from these distributions. This gives our model the flexibility to approach ZSL from a different perspective by treating it as a missing data problem. More specifically, once the model parameters $\vect{\Theta}$ and $\vect{\Phi}$ are trained, we can obtain the distribution embedding of any unseen class by just passing its corresponding class-attribute vector through the mapping network $f_{\vect{\Phi}}$. As a result, one can synthesize examples of seen as well as unseen classes in the latent space from their respective distribution embeddings and use them to train a supervised classifier. Following this perspective, a few methods~\cite{mishra2018generative, chen2018zero,schonfeld2019generalized} have shown a significant improvement over methods that do not use synthetic examples, especially in the GZSL setting. This can be explained by the fact that only examples from the seen classes are often used to learn a ZSL model, therefore, it can produce biases towards the seen classes~\cite{chao2016empirical}. To further validate this claim, we will evaluate our model in the GZSL setting using a standard nearest-neighbor classifier without synthetic examples and also a linear softmax classifier trained with the synthetic examples.


\subsection{Implementation Details}
The mapping neural networks $f_{\vect{\Theta}}$ and $f_{\vect{\Phi}}$ are modeled as MLPs with only one hidden layer, followed by two output layers representing a mean and a diagonal covariance matrix of a Gaussian distribution. To ensure the positive semi-definiteness of the covariance matrix, we model the logarithm of the diagonal covariance matrix as a $K$-dimensional real vector. Each hidden layer is followed by a batch normalization layer and a dropout layer. We set a dropout rate of 0.5 for $f_{\vect{\Theta}}$ and 0.1 for $f_{\vect{\Phi}}$. Rectified Linear Unit~\cite{icmlrelu} (ReLU) is used as activation functions for the hidden layer. Our model is trained with stochastic gradient descent using the Adam optimizer~\cite{kingma2014adam} with a mini-batch size of 64 and a learning rate of $10^{-5}$. The margin for the hinge loss function in Eq.~(\ref{eq:batchloss}) is set to $\epsilon=1$ for all experiments. For the purpose of reproducibility, all source codes of our method will be available.


\section{Experiments}
In this section, we evaluate the effectiveness of the proposed method on multiple ZSL data sets. An experimental comparison with other state-of-the-art methods is provided for both the ZSL and GZSL settings. For further analysis, we carry extensive ablation studies to show the effectiveness of the proposal.

\subsection{Data Sets and Settings}
We conduct extensive experiments on three widely-used benchmark data sets. A brief description of these data sets is given as follows. Animals with Attributes (\textbf{AWA2})~\cite{xian2018zero} is one of the most popular data sets to evaluate a ZSL model. It contains 37,322 images of animals from 50 classes with diverse backgrounds. The data set includes an 85-dimensional human-annotated-attribute vector for each class. These attributes contain both binary and continuous features. For ZSL, we use 40 classes to train/validation and 10 classes to test. The SUN Scene Recognition (\textbf{SUN})~\cite{xiao2010sun} data set consists of 14,340 images from 717 different scenes for fine-grained image classification. Each class has a limited number of images, making it very challenging for ZSL. Since each image is accompanied with a 102-dimensional human-annotated-attribute vector, we average over such vectors from images of the same class to get a unique class-attribute vector representation. This data set is split into 645 seen and 72 unseen classes. Caltech-UCSD Birds 200 (\textbf{CUB})~\cite{WelinderEtal2010} is another fine-grained data set containing 11,788 images from 200 classes of different bird species. Each class contains approximately 60 images. To make it suitable for ZSL, we use a conventional split to divide the bird species into 150 seen classes and 50 unseen classes. Following Verma \etal~\cite{verma2020meta}, we use the character-level convolutional recurrent neural textual (CRNN) features~\cite{reed2016learning} to obtain a 1024-dimensional vector as class attributes for each class label. In order to achieve a fair comparison, all data splits for training and test sets follow the standard evaluation protocol proposed by Xian \etal~\cite{xian2018zero}. We use the 2048-dimensional ResNet-101~\cite{he2016deep} features, which are pre-trained on ImageNet with 1,000 classes, for all the images. Note that the data splits ensure that none of the test classes appear in the 1,000 classes used to train ResNet-101. 


As commonly done, we report the average per-class top-1 accuracy as an evaluation metric for the traditional ZSL setting. This is an important evaluation metric, especially when the data set is not well balanced as in ZSL.  In the GZSL setting, we use the harmonic mean as proposed by Xian \etal~\cite{xian2018zero}, given by
\begin{align}
    H = \frac{2*S*U}{S+U}\,,
\end{align}
where $S$ and $U$ denote the average per-class top-1 accuracy over the seen and unseen classes, respectively. The reason for using the harmonic mean instead of the arithmetic mean is because the latter may induce biases when the accuracy over the seen classes is much higher than that of unseen classes. Consistent ZSL methods should achieve high accuracy on both seen as well as unseen classes.


\subsection{Competing Methods}
We compare our method DETC with several state-of-the-art methods, which use the same data splits. ESZSL~\cite{romera2015embarrassingly}, SAE~\cite{kodirov2017semantic}, SYNC~\cite{changpinyo2016synthesized}, LATEM~\cite{xian2016latent}, and DeViSE~\cite{frome2013devise} are mapping-based methods that learn a compatibility function between the image features and class attributes. Similarly to our method, GFZSL~\cite{verma2017simple} and ZSL-ADA~\cite{khare2020generative} also model the seen and unseen class labels as distributions, but they use a different training paradigm. E-PGN~\cite{yu2020episode} uses a much more complicated training framework for ZSL where training process is divided into extensive episodes and each episode mimics a ZSL task. SJE~\cite{akata2015evaluation} and SP-AEN~\cite{chen2018zero} are based on a large-margin framework. SE-ZSL~\cite{kumar2018generalized}, CVAE~\cite{mishra2018generative}, and f-CLSWGAN~\cite{xian2018feature} treat ZSL as a missing data problem and synthesize examples from unseen classes. ReViSE~\cite{hubert2017learning} learns a shared latent space between the image features and class attributes using autoencoders, while CADA-VAE~\cite{schonfeld2019generalized} is based on aligned variational autoencoders to learn the latent space. For a fair comparison,  all the competing methods are evaluated on the same image features. The results of GFZSL, ZSL-ADA, and CADA-VAE are taken from their original publications, while the rest of the results are taken from Verma \etal~\cite{verma2020meta}, where they reported under the same settings like ours.

\subsection{Zero-Shot Learning}
In this subsection, we provide some insights about the performance of the competing methods under the traditional ZSL setting, in which the test data only contain unseen classes. The experimental results are shown in Table~\ref{ZSL_Results}. Despite its simplicity, the proposed method DETC outperforms most of the competing methods. Especially, it obtains a relative improvement of $4.6\%$ compared to the strongest competitor on the CUB data set, while obtaining competitive results on the SUN and AWA2 data sets. As expected, our distribution embedding-based method is more powerful than the vector embedding-based methods (\emph{e.g.}, DeViSE, LATEM, SAE, SYNC, and SJE). Another observation is that there is a big performance gap between our model and other deep learning-based methods such as SP-AEN, f-CLSWGAN, and SE-ZSL. DETC outperforms not only simple generative frameworks like GFZSL, ZSL-ADA, but also more complex generative models like f-CLSWGAN, CADA-VAE, and CVAE.  This is because the triplet constraints are explicitly used in our method to preserve the similarity relationship between image features and class attributes. As a result, DETC can generalize very well on unseen classes. Moreover, compared to other complex generative models (\emph{e.g.}, CVAE, CADA-VAE and f-CLSWGAN), DETC is much simpler to implement.


\begin{table}[!t]
\centering
\begin{tabular}{|l||l|l|l|}
\hline
\textbf{Method}  & \textbf{AWA2} & \textbf{CUB}  & \textbf{SUN}  \\ \hline \hline
SJE              & 61.9          & 53.9          & 53.7          \\
ESZSL            & 58.6          & 53.9          & 54.5          \\
SYNC          & 46.6          & 55.6          & 56.3          \\
SAE              & 54.1          & 33.3          & 40.3          \\
LATEM            & 55.8          & 49.3          & 55.3          \\
DeViSE           & 59.7          & 52.0          & 56.5          \\
GFZSL            & 67.0          & 49.2          & 60.6          \\
CVAE         & 65.8          & 52.1          & 61.7          \\
SE-ZSL           & 69.2          & 59.6          & \textbf{63.4}$^{*}$ \\
f-CLSWGAN        & 68.2          & 57.3          & 60.8          \\
SP-AEN           & 58.5          & 55.4          & -             \\
CADA-VAE         & 64.0          & 64.6          & 61.8          \\
ZSL-ADA          & \textbf{70.4} & 70.9          & 63.3 \\ 
E-PGN & \textbf{73.4}$^{*}$ & \textbf{72.4} & -\\
\hline \hline
DETC (Ours)          & 70.0 & \textbf{74.6}$^{*}$ & \textbf{64.2}$^{*}$ \\
\hline        
\end{tabular}
\caption{Zero-shot learning results on the AWA2, CUB, and SUN data sets. We measure the average per-class top-1 accuracy in \%. The best results are in bold with an asterisk and the second best results are in bold. }
\label{ZSL_Results}
\end{table}

\subsection{Generalized Zero-Short Learning}\label{GZSL}
Real applications often come with more realistic conditions where we do not know whether a test image belongs to a seen or unseen class. Therefore, GZSL can provide a more practical point of view to assess a ZSL method. In this subsection, we aim to evaluate the performance of the competing methods on both seen and unseen classes. The same data splits as in the ZSL setting are used. Note that we have a separate test set for the seen classes for each data set. These test sets are not used during training. The experimental results are shown in Table~\ref{GZSL_Results}. The overall performance drop is due to the increasing number of classes on the test sets. For DETC, we report the GZSL results using a nearest-neighbor classifier (see the penultimate row in Table~\ref{GZSL_Results}). To achieve a balance between examples from seen and unseen classes, DETC can generate low-dimensional latent features, on which we train a softmax classifier. A remarkable improvement is obtained (see the last row in Table~\ref{GZSL_Results}). This improvement is attributed to the synthetic examples, which help to reduce the seen class bias significantly. A large improvement can be observed for the SUN data set, which contains 645 fine-grained seen classes. For the AWA2 and CUB data sets, our method yields the best harmonic mean accuracy and the best average per-class top-1 accuracy on the unseen classes. It achieves a relative improvement of $8.2\%$ for CUB and $0.6\%$ for AWA2 compared to the strongest competitor. Besides, our method obtains a competitive result on the SUN data set. This superior performance again confirms the robustness of distribution embeddings learned by our method. It is noted that SYNC obtains a very high average per-class top-1 accuracy on the seen classes, however, it fails to distinguish between the unseen classes. This is obviously very undesirable in ZSL.

\begin{table*}
\centering
\begin{tabular}{|l|lll|lll|lll|}
\hline
\multicolumn{1}{|c|}{\multirow{2}{*}{\textbf{Method}}} & \multicolumn{3}{c|}{\textbf{AWA2}}                                       & \multicolumn{3}{c|}{\textbf{CUB}}                                                 & \multicolumn{3}{c|}{\textbf{SUN}}                                                 \\ \cline{2-10}
\multicolumn{1}{|c|}{}                                 & \multicolumn{1}{c|}{U} & \multicolumn{1}{c|}{S} & \multicolumn{1}{c|}{H} & \multicolumn{1}{c|}{U} & \multicolumn{1}{c|}{S} & \multicolumn{1}{c|}{H} & \multicolumn{1}{c|}{U} & \multicolumn{1}{c|}{S} & \multicolumn{1}{c|}{H} \\ \hline \hline
SJE                                                    & 8.0                    & 73.9                   & 14.4                   & 23.5                   & 59.2                   & 33.6                   & 14.7                   & 30.5                   & 19.8                   \\
ESZSL                                                  & 5.9                    & 77.8                   & 11.0                   & 12.6                   & 63.8                   & 21.0                   & 11.0                   & 27.9                   & 15.8                   \\
SAE                                                    & 1.1                    & 82.2                   & 2.2                    & 7.8                    & 54.0                   & 13.6                   & 8.8                    & 18.0                   & 11.8                   \\
SYNC                                                   & 10.0                   & 90.5                   & 18.0                   & 11.5                   & 70.9                   & 19.8                   & 7.9                    & 43.3                   & 13.4                   \\
LATEM                                                  & 11.5                   & 77.3                   & 20.0                   & 15.2                   & 57.3                   & 24.0                   & 14.7                   & 28.8                   & 19.5                   \\
DeViSE                                                 & 17.1                   & 74.7                   & 27.8                   & 23.                    & 53.0                   & 32.8                   & 16.9                   & 27.4                   & 20.9                   \\
ReViSE                                                 & 39.7                   & 46.4                   & 42.8                   & 28.3                   & 37.6                   & 32.3                   & 20.1                   & 24.3                   & 22.0                   \\
CVAE                                                   & -                      & -                      & 51.2                   & -                      & -                      & 34.5                   & -                      & -                      & 26.7                   \\
SE-ZSL                                                 & 58.3                   & 68.1                   & 62.8                   & 41.5                   & 53.3                   & 46.7                   & 40.9                   & 30.5                   & 34.9                   \\
f-CLSWGAN                                              & 57.9                   & 61.4                   & 59.6                   & 43.7                   & 57.7                   & 49.7                   & 42.6                   & 36.6                   & \textbf{39.4}                   \\
CADA-VAE                                               & 55.8                   & 75.0                   & 63.9                   & 51.6                   & 55.5                   & 53.4                   & 47.2                   & 35.7                   & \textbf{40.6}$^{*}$ \\
E-PGN & 52.6    & 83.5  & \textbf{64.6}$^{*}$  & 52.0  & 61.1    & 56.2  & - & - & -
\\ \hline \hline
DETC (Ours) & 53.9 & 74.4  & 62.5           & 49.2 & 68.6 & \textbf{57.3}            & 20.8  & 37.9  & 26.9 \\
DETC* (Ours) & 58.3 & 71.6  & \textbf{64.3}         & 60.6 & 58.5  & \textbf{59.5}$^{*}$                   & 42.4   & 36.4   & 39.2 \\ \hline                 
\end{tabular}
\caption{GZSL results on the AWA2, CUB, and SUN data sets. Here, U and S denote the average per-class top-1 accuracy over unseen and seen classes, respectively. The harmonic mean of U and S is denoted by H. The penultimate row shows the performance of DETC using a nearest-neighbor classifier, while the last row shows its performance using a linear softmax classifier. The best results are in bold with an asterisk and the second-best results are in bold.}
\label{GZSL_Results}
\end{table*}


\begin{table*}
\centering
\begin{tabular}{|l|lll|lll|lll|}
\hline
\multicolumn{1}{|c|}{\multirow{2}{*}{\textbf{Method}}} & \multicolumn{3}{c|}{\textbf{AWA2}}                                       & \multicolumn{3}{c|}{\textbf{CUB}}                                                 & \multicolumn{3}{c|}{\textbf{SUN}}                                                 \\ \cline{2-10}
\multicolumn{1}{|c|}{}                                 & \multicolumn{1}{c|}{U} & \multicolumn{1}{c|}{S} & \multicolumn{1}{c|}{H} & \multicolumn{1}{c|}{U} & \multicolumn{1}{c|}{S} & \multicolumn{1}{c|}{H} & \multicolumn{1}{c|}{U} & \multicolumn{1}{c|}{S} & \multicolumn{1}{c|}{H} \\ \hline \hline
Vector Embeddings	& 50.2 & 71.6 & 59.0 & 47.8 & 67.2 & 55.8 & 17.2 & \textbf{39.9} & 24.0 \\
Distribution Embeddings	& \textbf{53.9} & \textbf{74.4} & \textbf{62.5} & \textbf{50.0} & \textbf{67.7} & \textbf{57.5} & \textbf{20.8} & 37.9 & \textbf{26.9} 
 \\ \hline                 
\end{tabular}
\caption{GZSL results on the AWA2, CUB and SUN data sets of DETC using vector embeddings and distributions embeddings.}
\label{vector_vs_distribution_Results}
\end{table*}

\subsection{Further Analysis}
In this subsection, we further perform ablation studies to analyze the sensitivity of DETC and the effects of data generation in the GZSL setting.

\subsubsection{Distribution Embeddings vs Vector Embeddings}
In this experiment, we provide empirical evidence for the effectiveness of distribution embeddings over vector embeddings. To this end, instead of using the 2-Wasserstein distance metric between distributions, we simply use the Euclidean distance metric between the mean vectors to train DETC in GZSL settings. For a fair comparison, we perform a nearest-neighbor search to report the classification results. Table~\ref{vector_vs_distribution_Results} summarizes the results of this experiment. We can clearly see that DETC with distribution embeddings shows a clear improvement over vector embeddings.  Results from distribution embeddings can further be improved by data generation as explained in Section~\ref{GZSL}.

\subsubsection{Visualization}
\begin{figure}[!t]
    \centering
    \includegraphics[width=0.5\textwidth]{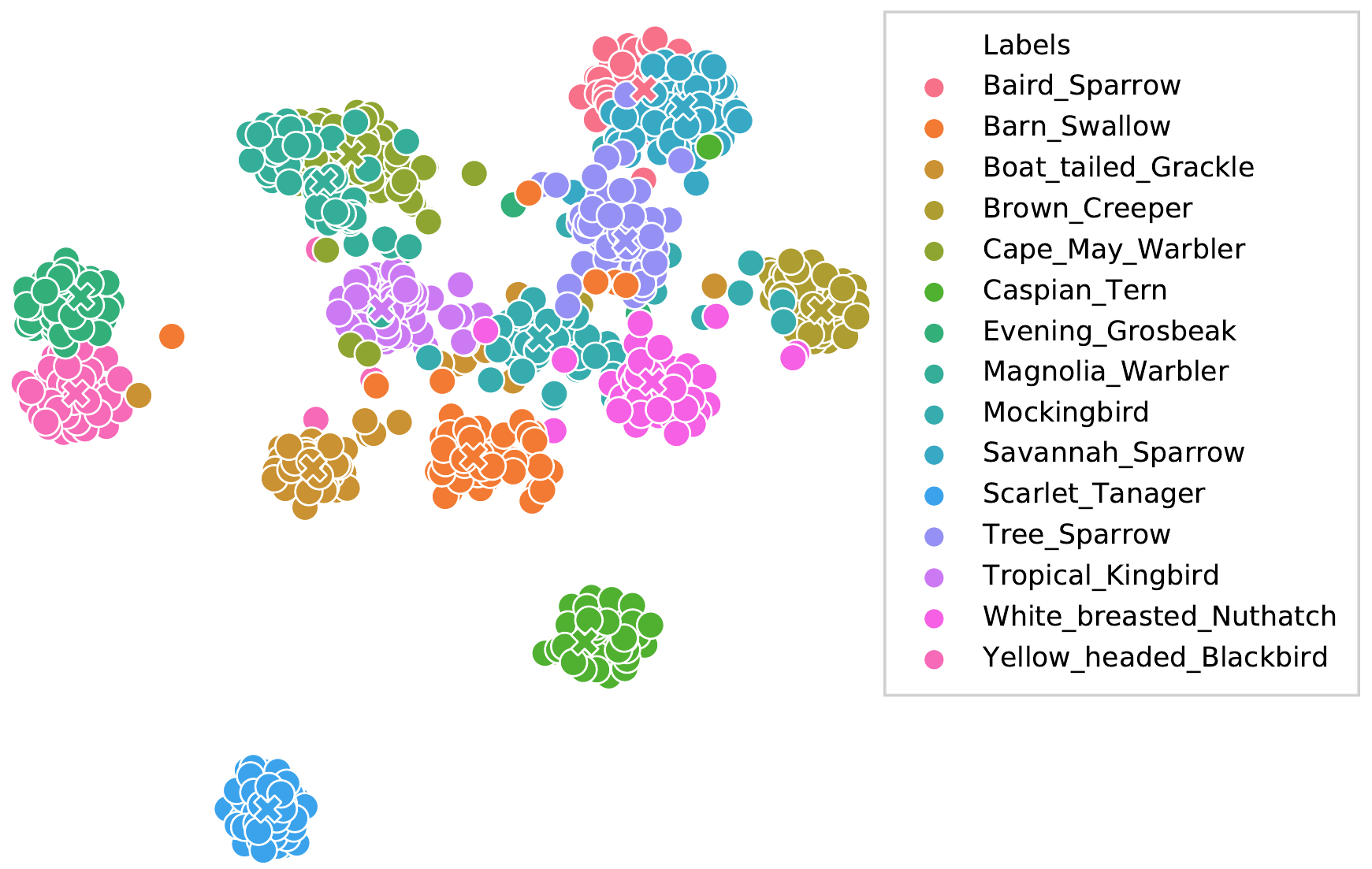}
  \caption{The Barnes-Hut t-SNE visualization of the feature embeddings computed by DETC on a subset of the CUB data set. Each point denotes the mean of its corresponding distribution. The class-attribute means are also shown with $\times$ mark.}
  \label{fig:CUB_tsne}
\end{figure}

The Barnes-Hut t-SNE~\cite{JMLR:v15:vandermaaten14a} visualization on the CUB data set is shown in Fig.~\ref{fig:CUB_tsne}. For clarity, we use examples from 15 randomly chosen unseen classes. Only the means of the distributions in the latent spaces are plotted. As shown in this figure, examples of the same class are clustered together around their correct class embedding, whereas examples of different classes are pushed far apart from each other.

\subsubsection{Analysis of Dissimilarity Measures}
\begin{figure}[!t]
    \centering
    \includegraphics[width=0.5\textwidth]{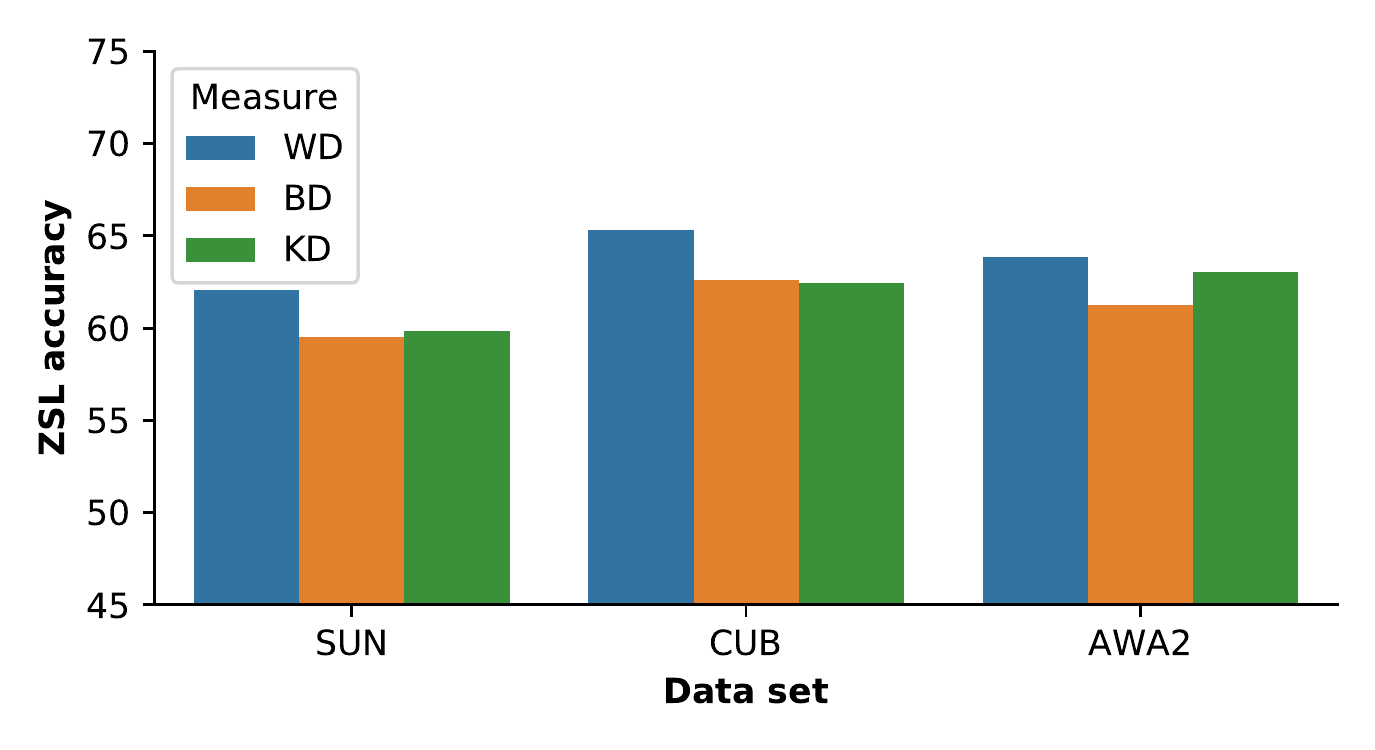}
    \caption{Average per-class top-1 accuracy of DETC using different dissimilarity measures on the AWA2, CUB, and SUN data sets.}
    \label{fig:Dist_eval_fig}
\end{figure}

In this experiment, we evaluate the performance of DETC using different dissimilarity measures to compute the distance between two distributions. To this end, we replace the Wasserstein distance (\textbf{WD}) used in problem~(\ref{eq:batchloss}) by another common distance measure, including the Kullback–Leibler divergence (\textbf{KD}) and the Bhattacharya distance (\textbf{BD})~\cite{bhattacharyya1943measure}. Although the Kullback–Leibler divergence measures the dissimilarity between two distribution, it is not a distance metric due to the asymmetry. The same experimental settings as in the previous subsection are configured. The results are presented in Fig.~\ref{fig:Dist_eval_fig}. As we can see from the figure, the proposed method DETC using WD clearly outperforms other dissimilarity measures over all the data sets.


\subsubsection{Effects of Data Generation} 
\begin{figure}[!t]
    \centering
    \includegraphics[width=0.35\textwidth]{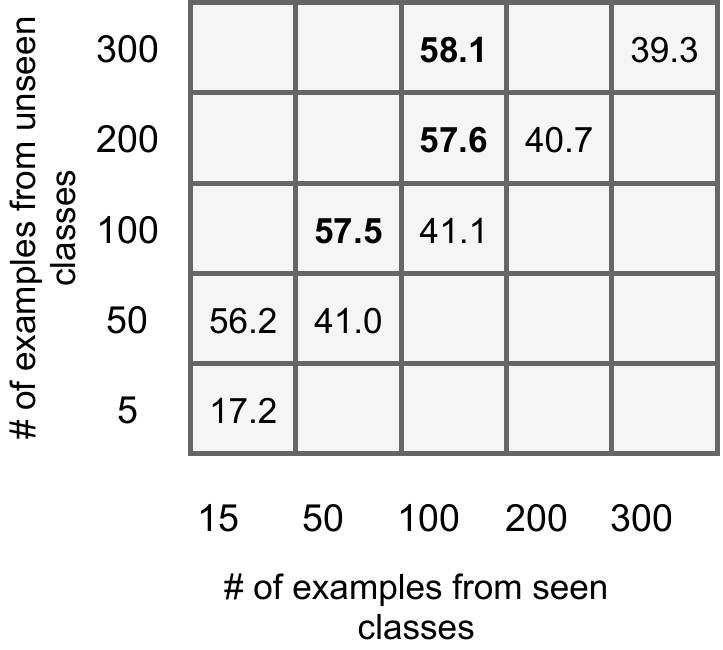}
  \caption{Analyzing the effects of the ratio between the number of synthetic examples from the seen and unseen classes on the harmonic mean accuracy in GZSL for the CUB data set.}
  \label{GZSL_Sampling}
\end{figure}

Given a class-attribute vector, we can obtain its distribution embedding in the latent space. This distribution can be used to generate examples in the latent space. The GZSL performances of our method using different ratios of examples sampled from the seen and unseen classes are reported in Table~\ref{GZSL_Sampling}. As can be observed from the table, these ratios have an important influence on the overall GZSL performance. The best harmonic mean accuracy is achieved with the ratio of 1:2 or 1:3 between the number of examples from the seen and unseen classes.

\section{Conclusions}
In this paper, we have proposed the use of distribution embeddings for ZSL. In contrast to the existing vector-embedding-based methods, both image features and class attributes are represented as distributions in a latent space, leading to a better modeling of the intra-class variability. Our method is based on large-margin learning constraints with the goal of minimizing the Wasserstein distances of an image to its correct label and maximizing the distances to its incorrect labels. Learning is formulated in an end-to-end manner. Despite the simple philosophy behind our proposed method, it achieves the state-of-the-art performance on several benchmark data sets for both the ZSL and GZSL settings.

{\small
\bibliographystyle{ieee_fullname}
\bibliography{main}

\begin{thebibliography}{10}\itemsep=-1pt

\bibitem{akata2015label}
Zeynep Akata, Florent Perronnin, Zaid Harchaoui, and Cordelia Schmid.
\newblock Label-embedding for image classification.
\newblock {\em IEEE Transactions on Pattern Analysis and Machine Intelligence},
  38(7):1425--1438, 2015.

\bibitem{akata2015evaluation}
Zeynep Akata, Scott Reed, Daniel Walter, Honglak Lee, and Bernt Schiele.
\newblock Evaluation of output embeddings for fine-grained image
  classification.
\newblock In {\em Proceedings of the IEEE Conference on Computer Vision and
  Pattern Recognition}, pages 2927--2936, 2015.

\bibitem{pmlr-v70-arjovsky17a}
Martin Arjovsky, Soumith Chintala, and L{\'e}on Bottou.
\newblock {W}asserstein generative adversarial networks.
\newblock In {\em Proceedings of the International Conference on Machine
  Learning}, pages 214--223, 2017.

\bibitem{bhattacharyya1943measure}
Anil Bhattacharyya.
\newblock On a measure of divergence between two statistical populations
  defined by their probability distributions.
\newblock {\em Bull. Calcutta Math. Soc.}, 35:99--109, 1943.

\bibitem{bucher2016improving}
Maxime Bucher, St{\'e}phane Herbin, and Fr{\'e}d{\'e}ric Jurie.
\newblock Improving semantic embedding consistency by metric learning for
  zero-shot classiffication.
\newblock In {\em Proceedings of the European Conference on Computer Vision},
  pages 730--746. Springer, 2016.

\bibitem{changpinyo2016synthesized}
Soravit Changpinyo, Wei-Lun Chao, Boqing Gong, and Fei Sha.
\newblock Synthesized classifiers for zero-shot learning.
\newblock In {\em Proceedings of the IEEE conference on computer vision and
  pattern recognition}, pages 5327--5336, 2016.

\bibitem{chao2016empirical}
Wei-Lun Chao, Soravit Changpinyo, Boqing Gong, and Fei Sha.
\newblock An empirical study and analysis of generalized zero-shot learning for
  object recognition in the wild.
\newblock In {\em Proceedings of the European Conference on Computer Vision},
  pages 52--68, 2016.

\bibitem{chen2018zero}
Long Chen, Hanwang Zhang, Jun Xiao, Wei Liu, and Shih-Fu Chang.
\newblock Zero-shot visual recognition using semantics-preserving adversarial
  embedding networks.
\newblock In {\em Proceedings of the IEEE Conference on Computer Vision and
  Pattern Recognition}, pages 1043--1052, 2018.

\bibitem{Elhoseiny_2019_ICCV}
Mohamed Elhoseiny and Mohamed Elfeki.
\newblock Creativity inspired zero-shot learning.
\newblock In {\em Proceedings of the International Conference on Computer
  Vision}, 2019.

\bibitem{frome2013devise}
Andrea Frome, Greg~S Corrado, Jon Shlens, Samy Bengio, Jeff Dean, Marc'Aurelio
  Ranzato, and Tomas Mikolov.
\newblock Devise: A deep visual-semantic embedding model.
\newblock In {\em Advances in Neural Information Processing Systems}, pages
  2121--2129, 2013.

\bibitem{givens1984class}
Clark~R Givens, Rae~Michael Shortt, et~al.
\newblock A class of wasserstein metrics for probability distributions.
\newblock {\em The Michigan Mathematical Journal}, 31(2):231--240, 1984.

\bibitem{10.5555/3295222.3295327}
Ishaan Gulrajani, Faruk Ahmed, Martin Arjovsky, Vincent Dumoulin, and Aaron
  Courville.
\newblock Improved training of wasserstein gans.
\newblock In {\em Advances in Neural Information Processing Systems}, page
  5769–5779, 2017.

\bibitem{he2016deep}
Kaiming He, Xiangyu Zhang, Shaoqing Ren, and Jian Sun.
\newblock Deep residual learning for image recognition.
\newblock In {\em Proceedings of the IEEE Conference on Computer Vision and
  Pattern Recognition}, pages 770--778, 2016.

\bibitem{hubert2017learning}
Yao-Hung Hubert~Tsai, Liang-Kang Huang, and Ruslan Salakhutdinov.
\newblock Learning robust visual-semantic embeddings.
\newblock In {\em Proceedings of the IEEE International Conference on Computer
  Vision}, pages 3571--3580, 2017.

\bibitem{kankuekul2012online}
Pichai Kankuekul, Aram Kawewong, Sirinart Tangruamsub, and Osamu Hasegawa.
\newblock Online incremental attribute-based zero-shot learning.
\newblock In {\em Proceedings of the IEEE Conference on Computer Vision and
  Pattern Recognition}, pages 3657--3664, 2012.

\bibitem{karessli2017gaze}
Nour Karessli, Zeynep Akata, Bernt Schiele, and Andreas Bulling.
\newblock Gaze embeddings for zero-shot image classification.
\newblock In {\em Proceedings of the IEEE Conference on Computer Vision and
  Pattern Recognition}, pages 4525--4534, 2017.

\bibitem{Kato_2019_ICCV_Workshops}
Naoki Kato, Toshihiko Yamasaki, and Kiyoharu Aizawa.
\newblock Zero-shot semantic segmentation via variational mapping.
\newblock In {\em Proceedings of the International Conference on Computer
  Vision Workshops}, 2019.

\bibitem{kelley1960gradient}
Henry~J Kelley.
\newblock Gradient theory of optimal flight paths.
\newblock {\em Ars Journal}, 30(10):947--954, 1960.

\bibitem{khare2020generative}
Varun Khare, Divyat Mahajan, Homanga Bharadhwaj, Vinay~Kumar Verma, and Piyush
  Rai.
\newblock A generative framework for zero shot learning with adversarial domain
  adaptation.
\newblock In {\em Proceedings of the IEEE Winter Conference on Applications of
  Computer Vision}, pages 3101--3110, 2020.

\bibitem{kingma2014adam}
Diederik~P Kingma and Jimmy Ba.
\newblock Adam: A method for stochastic optimization.
\newblock In {\em Proceedings of the International Conference on Learning
  Representations}, 2014.

\bibitem{kodirov2017semantic}
Elyor Kodirov, Tao Xiang, and Shaogang Gong.
\newblock Semantic autoencoder for zero-shot learning.
\newblock In {\em Proceedings of the IEEE Conference on Computer Vision and
  Pattern Recognition}, pages 3174--3183, 2017.

\bibitem{kumar2018generalized}
Vinay Kumar~Verma, Gundeep Arora, Ashish Mishra, and Piyush Rai.
\newblock Generalized zero-shot learning via synthesized examples.
\newblock In {\em Proceedings of the IEEE Conference on Computer Vision and
  Pattern Recognition}, pages 4281--4289, 2018.

\bibitem{lampert2009learning}
Christoph~H Lampert, Hannes Nickisch, and Stefan Harmeling.
\newblock Learning to detect unseen object classes by between-class attribute
  transfer.
\newblock In {\em Proceedings of the IEEE Conference on Computer Vision and
  Pattern Recognition}, pages 951--958, 2009.

\bibitem{lampert2013attribute}
Christoph~H Lampert, Hannes Nickisch, and Stefan Harmeling.
\newblock Attribute-based classification for zero-shot visual object
  categorization.
\newblock {\em IEEE Transactions on Pattern Analysis and Machine Intelligence},
  36:453--465, 2013.

\bibitem{li2019zero}
Zhihui Li, Lina Yao, Xiaoqin Zhang, Xianzhi Wang, Salil Kanhere, and Huaxiang
  Zhang.
\newblock Zero-shot object detection with textual descriptions.
\newblock In {\em Proceedings of the AAAI Conference on Artificial
  Intelligence}, volume~33, pages 8690--8697, 2019.

\bibitem{miller1995wordnet}
George~A Miller.
\newblock Wordnet: a lexical database for english.
\newblock {\em Communications of the ACM}, 38:39--41, 1995.

\bibitem{mishra2018generative}
Ashish Mishra, Shiva Krishna~Reddy, Anurag Mittal, and Hema~A Murthy.
\newblock A generative model for zero shot learning using conditional
  variational autoencoders.
\newblock In {\em Proceedings of the IEEE Conference on Computer Vision and
  Pattern Recognition Workshops}, pages 2188--2196, 2018.

\bibitem{mukherjee2016gaussian}
Tanmoy Mukherjee and Timothy Hospedales.
\newblock Gaussian visual-linguistic embedding for zero-shot recognition.
\newblock In {\em Proceedings of the Conference on Empirical Methods in Natural
  Language Processing}, pages 912--918, 2016.

\bibitem{icmlrelu}
Vinod Nair and Geoffrey~E. Hinton.
\newblock Rectified linear units improve restricted boltzmann machines.
\newblock In {\em Proceedings of the International Conference on International
  Conference on Machine Learning}, page 807–814, 2010.

\bibitem{NGUYEN2020209}
Bac Nguyen and Bernard {De Baets}.
\newblock Improved deep embedding learning based on stochastic symmetric
  triplet loss and local sampling.
\newblock {\em Neurocomputing}, 402:209--219, 2020.

\bibitem{norouzi2013zero}
Mohammad Norouzi, Tomas Mikolov, Samy Bengio, Yoram Singer, Jonathon Shlens,
  Andrea Frome, Greg~S Corrado, and Jeffrey Dean.
\newblock Zero-shot learning by convex combination of semantic embeddings.
\newblock {\em arXiv preprint arXiv:1312.5650}, 2013.

\bibitem{palatucci2009zero}
Mark Palatucci, Dean Pomerleau, Geoffrey~E Hinton, and Tom~M Mitchell.
\newblock Zero-shot learning with semantic output codes.
\newblock In {\em Advances in neural information processing systems}, pages
  1410--1418, 2009.

\bibitem{pennington2014glove}
Jeffrey Pennington, Richard Socher, and Christopher~D Manning.
\newblock Glove: Global vectors for word representation.
\newblock In {\em Proceedings of the 2014 conference on empirical methods in
  natural language processing (EMNLP)}, pages 1532--1543, 2014.

\bibitem{reed2016learning}
Scott Reed, Zeynep Akata, Honglak Lee, and Bernt Schiele.
\newblock Learning deep representations of fine-grained visual descriptions.
\newblock In {\em Proceedings of the IEEE Conference on Computer Vision and
  Pattern Recognition}, pages 49--58, 2016.

\bibitem{romera2015embarrassingly}
Bernardino Romera-Paredes and Philip Torr.
\newblock An embarrassingly simple approach to zero-shot learning.
\newblock In {\em Proceedings of the International Conference on Machine
  Learning}, pages 2152--2161, 2015.

\bibitem{schonfeld2019generalized}
Edgar Schonfeld, Sayna Ebrahimi, Samarth Sinha, Trevor Darrell, and Zeynep
  Akata.
\newblock Generalized zero-and few-shot learning via aligned variational
  autoencoders.
\newblock In {\em Proceedings of the IEEE Conference on Computer Vision and
  Pattern Recognition}, pages 8247--8255, 2019.

\bibitem{simonyan2014very}
Karen Simonyan and Andrew Zisserman.
\newblock Very deep convolutional networks for large-scale image recognition.
\newblock {\em arXiv preprint arXiv:1409.1556}, 2014.

\bibitem{szegedy2015going}
Christian Szegedy, Wei Liu, Yangqing Jia, Pierre Sermanet, Scott Reed, Dragomir
  Anguelov, Dumitru Erhan, Vincent Vanhoucke, and Andrew Rabinovich.
\newblock Going deeper with convolutions.
\newblock In {\em Proceedings of the IEEE Conference on Computer Vision and
  Pattern Recognition}, pages 1--9, 2015.

\bibitem{JMLR:v15:vandermaaten14a}
Laurens van~der Maaten.
\newblock Accelerating t-{SNE} using tree-based algorithms.
\newblock {\em The Journal of Machine Learning Research}, 15:3221--3245, 2014.

\bibitem{verma2020meta}
Vinay~Kumar Verma, Dhanajit Brahma, and Piyush Rai.
\newblock Meta-learning for generalized zero-shot learning.
\newblock In {\em Proceedings of the AAAI Conference on Artificial
  Intelligence}, pages 6062--6069, 2020.

\bibitem{verma2017simple}
Vinay~Kumar Verma and Piyush Rai.
\newblock A simple exponential family framework for zero-shot learning.
\newblock In {\em Proceedings of the Joint European Conference on Machine
  Learning and Knowledge Discovery in Databases}, pages 792--808, 2017.

\bibitem{wang2019survey}
Wei Wang, Vincent~W Zheng, Han Yu, and Chunyan Miao.
\newblock A survey of zero-shot learning: Settings, methods, and applications.
\newblock {\em ACM Transactions on Intelligent Systems and Technology (TIST)},
  10(2):1--37, 2019.

\bibitem{WelinderEtal2010}
P. Welinder, S. Branson, T. Mita, C. Wah, F. Schroff, S. Belongie, and P.
  Perona.
\newblock {Caltech-UCSD Birds 200}.
\newblock Technical Report CNS-TR-2010-001, California Institute of Technology,
  2010.

\bibitem{xian2016latent}
Yongqin Xian, Zeynep Akata, Gaurav Sharma, Quynh Nguyen, Matthias Hein, and
  Bernt Schiele.
\newblock Latent embeddings for zero-shot classification.
\newblock In {\em Proceedings of the IEEE Conference on Computer Vision and
  Pattern Recognition}, pages 69--77, 2016.

\bibitem{xian2018zero}
Yongqin Xian, Christoph~H Lampert, Bernt Schiele, and Zeynep Akata.
\newblock Zero-shot learning—a comprehensive evaluation of the good, the bad
  and the ugly.
\newblock {\em IEEE Transactions on Pattern Analysis and Machine Intelligence},
  41:2251--2265, 2018.

\bibitem{xian2018feature}
Yongqin Xian, Tobias Lorenz, Bernt Schiele, and Zeynep Akata.
\newblock Feature generating networks for zero-shot learning.
\newblock In {\em Proceedings of the IEEE Conference on Computer Vision and
  Pattern Recognition}, pages 5542--5551, 2018.

\bibitem{xiao2010sun}
Jianxiong Xiao, James Hays, Krista~A Ehinger, Aude Oliva, and Antonio Torralba.
\newblock Sun database: Large-scale scene recognition from abbey to zoo.
\newblock In {\em Proceedings of the IEEE conference on computer vision and
  pattern recognition}, pages 3485--3492, 2010.

\bibitem{yu2020episode}
Yunlong Yu, Zhong Ji, Jungong Han, and Zhongfei Zhang.
\newblock Episode-based prototype generating network for zero-shot learning.
\newblock In {\em Proceedings of the IEEE/CVF Conference on Computer Vision and
  Pattern Recognition}, pages 14035--14044, 2020.

\bibitem{9051798}
L. {Zhang}, P. {Wang}, L. {Liu}, C. {Shen}, W. {Wei}, Y. {Zhang}, and A. {Van
  Den Hengel}.
\newblock Towards effective deep embedding for zero-shot learning.
\newblock {\em IEEE Transactions on Circuits and Systems for Video Technology},
  30:2843 -- 2852, 2020.

\bibitem{zhang2017learning}
Li Zhang, Tao Xiang, and Shaogang Gong.
\newblock Learning a deep embedding model for zero-shot learning.
\newblock In {\em Proceedings of the IEEE Conference on Computer Vision and
  Pattern Recognition}, pages 2021--2030, 2017.

\bibitem{zhu2018generative}
Yizhe Zhu, Mohamed Elhoseiny, Bingchen Liu, Xi Peng, and Ahmed Elgammal.
\newblock A generative adversarial approach for zero-shot learning from noisy
  texts.
\newblock In {\em Proceedings of the IEEE Conference on Computer Vision and
  Pattern Recognition}, pages 1004--1013, 2018.

\end{thebibliography}
}

\end{document}